\documentclass{article}

\usepackage[final]{neurips_2019}

\usepackage[utf8]{inputenc}
\usepackage[T1]{fontenc}
\usepackage{hyperref}
\usepackage{url}
\usepackage{booktabs}
\usepackage{amsfonts}
\usepackage{amsmath}
\usepackage{nicefrac}
\usepackage{microtype}
\usepackage{graphicx}
\usepackage{caption}
\usepackage{subcaption}
\usepackage{xcolor}
\usepackage{lipsum}
\usepackage{multirow}
\usepackage[symbol]{footmisc}

\title{
  Multitask Fine-Tuning and Generative Adversarial Learning for Improved Auxiliary Classification
}

\author{
  Christopher Sun \\
  Department of Computer Science \\
  Stanford University \\
  \texttt{chrisun@stanford.edu} \\
  \And
  Abishek Satish \\
  Department of Computer Science \\
  Stanford University \\
  \texttt{absatish@stanford.edu}
}

\begin{document}

\maketitle

\begin{abstract}
In this study, we implement a novel BERT architecture for multitask fine-tuning on three downstream tasks: sentiment classification, paraphrase detection, and semantic textual similarity prediction. Our model, Multitask BERT, incorporates layer sharing and a triplet architecture, custom sentence pair tokenization, loss pairing, and gradient surgery. Such optimizations yield a 0.516 sentiment classification accuracy, 0.886 paraphase detection accuracy, and 0.864 semantic textual similarity correlation on test data. We also apply generative adversarial learning to BERT, constructing a conditional generator model that maps from latent space to create fake embeddings in $\mathbb{R}^{768}$. These fake embeddings are concatenated with real BERT embeddings and passed into a discriminator model for auxiliary classification. Using this framework, which we refer to as AC-GAN-BERT, we conduct semi-supervised sensitivity analyses to investigate the effect of increasing amounts of unlabeled training data on AC-GAN-BERT's test accuracy. Overall, aside from implementing a high-performing multitask classification system, our novelty lies in the application of adversarial learning to construct a generator that mimics BERT. We find that the conditional generator successfully produces rich embeddings with clear spatial correlation with class labels, demonstrating avoidance of mode collapse. Our findings validate the GAN-BERT approach and point to future directions of generator-aided knowledge distillation.
\end{abstract}

\section{Introduction and Related Work}
Bidirectional Encoder Representations from Transformers (BERT), a pretrained transformer-based model, has revolutionized the field of Natural Language Processing (NLP) with its ability to produce robust contextual word representations, which have in turn been used to fine-tune models and achieve state-of-the-art performance in various NLP tasks \cite{devlin2018bert}\cite{tenney2019bert}\cite{liu2019multi}. Because of BERT's notable use of self-attention, the model can contextually learn and understand each word's importance to a sentence. 

In this work, we apply BERT embeddings and multitask learning to three downstream tasks: sentiment classification, paraphrase detection, and semantic textual similarity prediction. Multitask learning aims to enhance model performance by using shared representations to improve a classifier's generalization ability and alleviate challenges of task interference and imbalanced learning priorities. Liu et al. (2019) describes the complications that follow with multitask learning; when constructing their Multi-Task Deep Neural Network (MT-DNN), the authors emphasize the importance of finding cross-task data and using regularization to fight probable overfitting \cite{liu2019multi}. Our multitask BERT model takes inspiration from Bi et al.'s summation of loss functions across tasks for gradient calculations and parameter updates. However, such additive losses succumb to the issue of destructive gradients \cite{yu2020gradient}. Yu et al. (2020) propose a solution called gradient surgery, a technique that projects the gradient of a specific task onto the normal plane of another task's conflicting gradient. This method, called PCGrad, helps alleviate the destructive gradient problem \cite{yu2020gradient}.

The goal of this work is to propose a high-performing BERT-based architecture for the aforementioned NLP tasks. We first implement several optimizations for fine-tuning, including data augmentation, architectural modifications, hyperparameter tuning, and adjustments in loss and gradient calculation. We then apply generative adversarial learning to BERT, improving the GAN-BERT framework developed by Croce et al. (2020), in which a discriminator conducts a $k+1$ classification problem using both real BERT embeddings and fake generator embeddings \cite{croce2020gan}. GAN-BERT has shown promise for the incorporation of largely-unlabeled training data in semi-supervised training \cite{croce2020gan}. Our GAN methodology comprises a significant part of our work going beyond the default final project.

\section{Approach}
\subsection{Multitask BERT}
\begin{figure}[hbt!]
     \centering
     \includegraphics[scale=0.38]{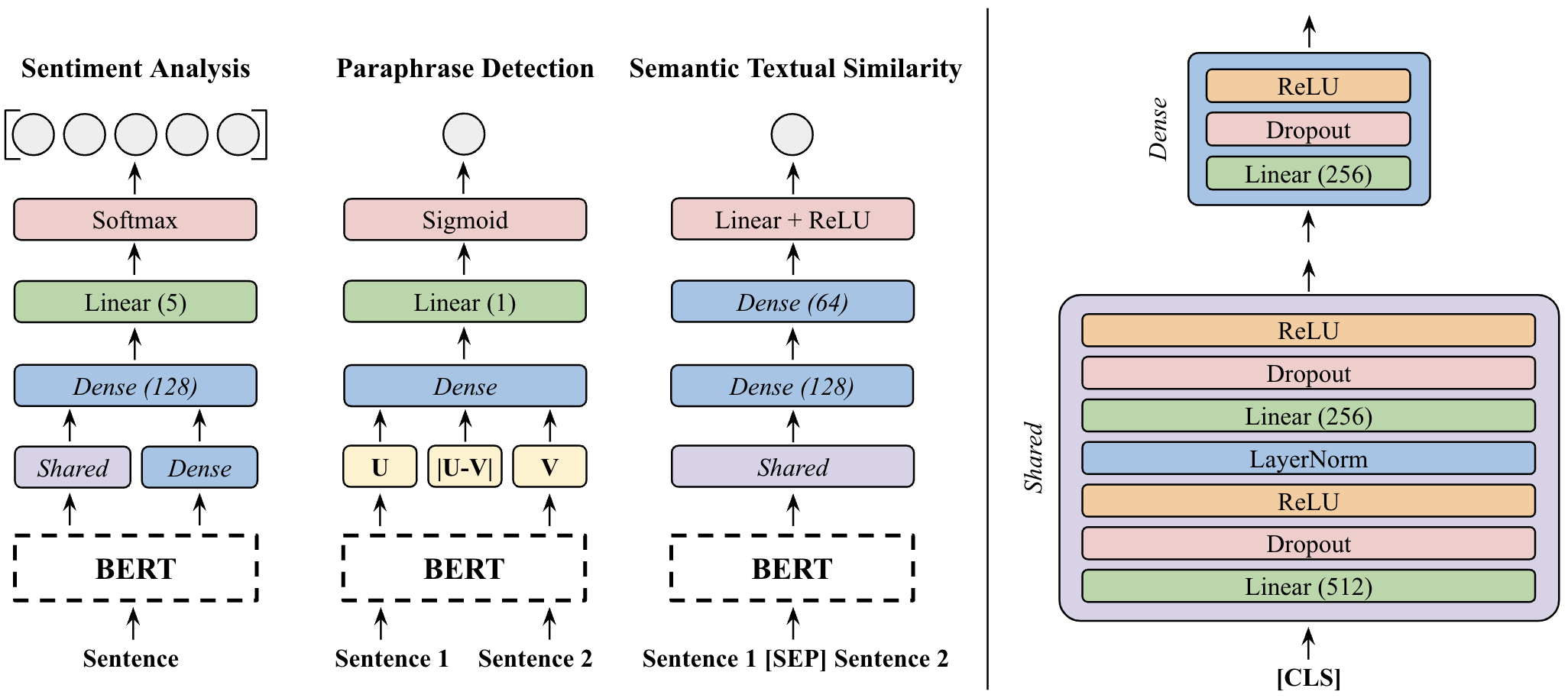}
     \caption{Multitask BERT Architecture}
     \label{fig:multitask_bert}
\end{figure}

We apply a suite of optimizations to improve BERT's performance on downstream NLP tasks, including multitask fine-tuning, manipulation of loss and gradient computations, and external augmentation of training data. After implementing the base version of BERT (minBERT) and loading pretrained weights using the provided starter code, we devise a novel head architecture for the simultaneous completion of our three NLP tasks of interest. Figure \ref{fig:multitask_bert} depicts the flow of information through our classification system: for each prediction head, we pass raw sentence(s) into BERT, extracting the special classification token $\texttt{[CLS]}\in\mathbb{R}^H$, where $H$ is the the hidden dimension of 768. We use layer sharing as motivated by previous success of MT-DNN in Liu et al. (2019) and biomedical multitask learning in Peng et al. (2020) \cite{liu2019multi} \cite{peng2020empirical}. In our \textit{shared} block, layer normalization is applied in between dense blocks because of training time benefits and batch size independence compared to batch normalization \cite{ba2016layer}. 
\paragraph{Sentiment Analysis} To classify phrase sentiment, we gather the \textit{shared} block output, simultaneously pass $\texttt{[CLS]}$ directly through a dense block, then concatenate these two representations and project them linearly. Softmax activation is applied to generate a probability distribution for multiclass classification.
\paragraph{Paraphrase Detection} We apply a triplet network to learn thematic similarity between paired sentences, as motivated by Dor et al. (2018) \cite{dor2018learning}. In Figure \ref{fig:multitask_bert}, $\mathbf{U}$ and $\mathbf{V}$ are alternate notation for respective $\texttt{[CLS]}$ embeddings retrieved from BERT, and $\lvert\mathbf{U-V}\rvert$ is the absolute difference in these embeddings. All three representations are concatenated and passed through dense layers, after which sigmoid activation is applied for binary classification.
\paragraph{Semantic Textual Similarity} Though the STS task also involves sentence pairs, we opt for a different tokenization mechanism: connecting two sentences as one with the $\texttt{[SEP]}$ token (we further discuss this decision in Section \ref{sec:results}). This combined embedding passes through the same \textit{shared block} as for sentiment analysis, as we reason logically that measures of semantic similarity align with sentiment relatedness.

Throughout the network, we apply dropout regularization to reduce overfitting \cite{srivastava2014dropout}.

\subsection{Multitask Fine-tuning Optimizations}
\label{sec:optimizations}
In multitask fine-tuning, a naive approach is to sum losses, then optimize parameters with respect to this summed loss, as in Bi et al. (2022) \cite{bi2022mtrec}. However, because of possible conflicting gradients between component losses for each prediction head, weights may be updated in suboptimal directions. To alleviate this issue, Yu et al. (2020) develops a gradient surgery technique called projecting conflicting gradients (PCGrad), which, simply put, involves modifying the gradient vectors before the parameter update step to ensure they are more aligned \cite{yu2020gradient}. In our approach, we optimize Multitask BERT with ``loss-pairing'' and gradient surgery: in our training loop, we apply PCGrad\footnote{We make use of Yu et al.'s source code linked in their paper.} twice by pairing losses as follows, where $SST$ corresponds to sentiment analysis, $P$ corresponds to paraphrase detection, and $STS$ corresponds to semantic similarity:
\[\mathcal{L}_1=[\mathcal{L}_{SST},\,\mathcal{L}_{P}],\,\mathcal{L}_2=[\mathcal{L}_{STS},\,\mathcal{L}_{P}].\]

The motivation for pairing losses as such is that we find paraphrase detection to be the highest-scoring, most smoothly-learning, most data-abundant task; taking this into account, we saw opportunity to use the gradient of $\mathcal{L}_P$ as the ``uniter'' between $\nabla\mathcal{L}_{SST}$ and $\nabla\mathcal{L}_{STS}$. Because of PCGrad, we can rest assured that having one stable and reliable task in each loss pair $\mathcal{L}_1$ and $\mathcal{L}_2$ will prevent even an outlandish gradient of $\mathcal{L}_{SST}$ or $\mathcal{L}_{STS}$ from throwing off the optimal update trajectory. We apply PCGrad on $\mathcal{L}_1$ and $\mathcal{L}_2$ sequentially, allowing the more optimal paraphrase learning to positively affect the gradient updates of $\mathcal{L}_{SST}$ and $\mathcal{L}_{STS}$ independently. Hence, we propose a combination of PCGrad and paraphrase-paired loss as the optimal way to fine-tune Multitask BERT. 

\subsection{Summary of Semi-Supervised GAN Training Framework}
\label{sec:gantraining}
\begin{figure}[hbt!]
     \centering
     \includegraphics[scale=0.3]{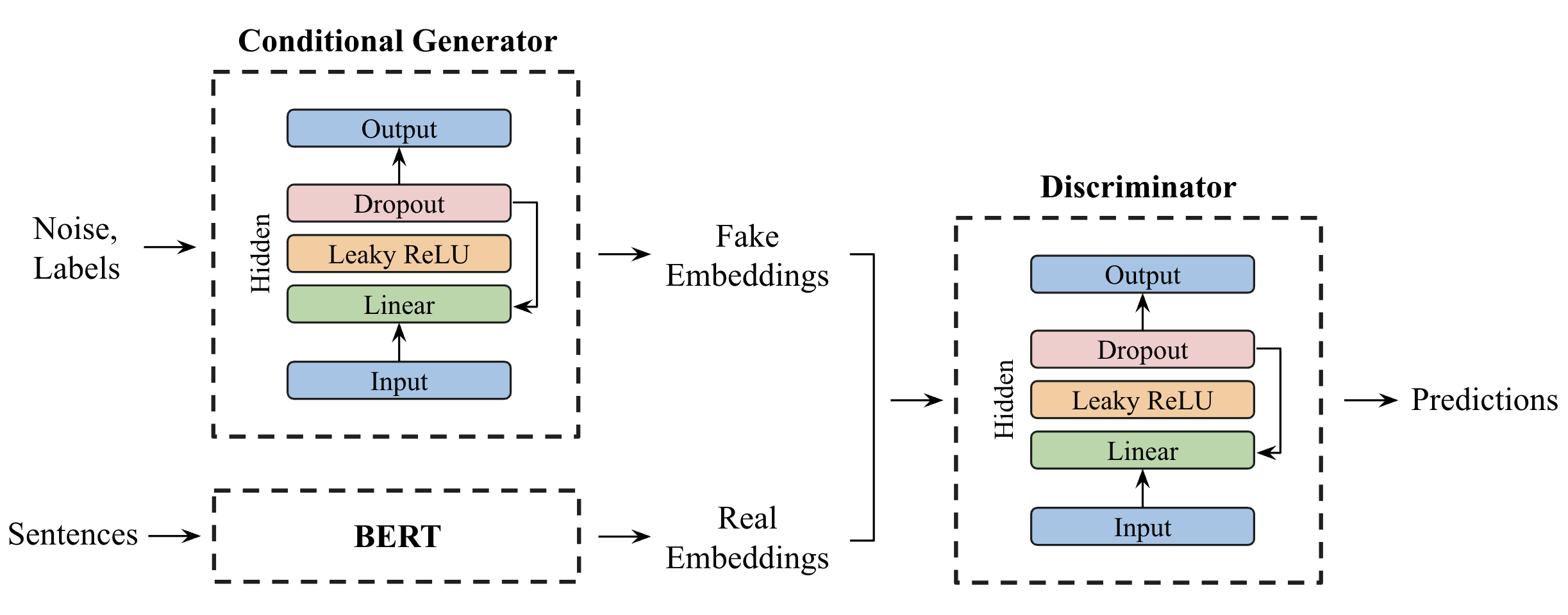}
     \caption{Auxiliary Classifier GAN-BERT Architecture (number of hidden layers variable)}
     \label{fig:ganbert}
\end{figure}

We validate the GAN-BERT\footnote{We inherit and make important modifications to Croce et al.'s source code.} framework and incorporate novel modifications, which we further assess in Section \ref{sec:analysis}. We develop a conditional Generative Adversarial Network (cGAN) from minBERT that acts as an auxiliary classifier for sentiment analysis and paraphrase detection. We refer to this model as AC-GAN-BERT, illustrated in Figure \ref{fig:ganbert}. 

The mathematical formulations in this section are directly adapted from Croce et al. (2020) \cite{croce2020gan}.
Let $p_d$ denote the real data distribution, $p_m$ denote model-predicted probabilities, and $p_G$ denote the generated examples.
\paragraph{Discriminator} The Discriminator $\mathcal{D}$ is a head for the BERT base, operating on both BERT $\texttt{[CLS]}$ embeddings and Generator (below) outputs for multiclass classification. The Discriminator loss $\mathcal{L_D}$ is a sum of supervised and unsupervised loss terms $\mathcal{L_{D_\text{S}}}$ and $\mathcal{L_{D_\text{U}}}$:
\begin{equation*}
    \begin{aligned}
        \mathcal{L_{D_\text{S}}}&=-\mathbb{E}_{x,y\sim p_d}\log[p_m(\hat{y}=y\vert x,y\in(1,\dots,k)] \\
        \mathcal{L_{D_\text{U}}}&=-\mathbb{E}_{x\sim p_d}\log[1-p_m(\hat{y}=y\vert x,y=k+1)]-\mathbb{E}_{x\sim\mathcal{G}}\log[p_m(\hat{y}=y\vert x,y=k+1)]
    \end{aligned}
\end{equation*}
In simple terms, $\mathcal{L_{D_\text{S}}}$ represents the misclassification of a real labeled example belonging to one of $k$ classes, and $\mathcal{L_{D_\text{U}}}$ represents the misclassification of an unlabeled example as real or fake.

\paragraph{Conditional Generator} The Conditional Generator $\mathcal{G}$ conditions on a class label to project from latent space to a fake embedding, trying to fool the Discriminator into thinking the embedding is real. The Generator loss $\mathcal{L_G}$ is a sum of ``feature matching'' and unsupervised loss terms $\mathcal{L_{G_\text{FM}}}$ and $\mathcal{L_{G_\text{U}}}$:
\begin{equation*}
    \begin{aligned}
        \mathcal{L_{G_\text{FM}}}&=\lVert \mathbb{E}_{x\sim p_d} f(x)-\mathbb{E}_{x\sim G} f(x)\rVert_2^2 \\
        \mathcal{L_{G_\text{U}}}&=-\mathbb{E}_{x\sim\mathcal{G}}\log[1-p_m(\hat{y}=y\vert x,y=k+1)]
    \end{aligned}
\end{equation*}
Since $\mathcal{G}$ ideally generates embeddings similar to $p_d$, $\mathcal{L_{G_\text{FM}}}$ examines the difference between an expected Generator example and the activation of an intermediate layer in $\mathcal{D}$ (hence the $f(x)$) expressed as a squared L2 norm. Finally, $\mathcal{L_{G_\text{U}}}$ considers the correct classification of fake (generated) embeddings by $\mathcal{D}$, which penalizes $\mathcal{G}$ because of the inability of $\mathcal{G}$ to fool $\mathcal{D}$.

\section{Experiments}
\subsection{Data}
\label{sec:data}
\paragraph{Sentiment Analysis} We use the Stanford Sentiment Treebank (SST) dataset for the sentiment classification task, which involves labeling phrases (215,154 in total) as negative, somewhat negative, neutral, somewhat positive, or positive \cite{socher2013recursive}. During fine-tuning, we also bring in the Massive Text Embedding Benchmark (MTEB) dataset consisting of 31,015 additional rows of data \cite{muennighoff2022mteb}. However, because MTEB labels range from 0-2, we scale them by 2; though this creates class imbalance, we find this additional data to be beneficial (further discussion in Section \ref{sec:analysis}.
\paragraph{Paraphrase Detection} We use the Quora Question Pairs dataset to predict a binary label representing whether questions pairs (404,298 in total) are paraphrases of each other \cite{quoradataset}.
\paragraph{Semantic Textual Similarity} We use the SemEval STS dataset for the regression task of predicting semantic similarity on a scale of 0 to 5 between two sentences (8,628 in total) \cite{agirre2013sem}. During fine-tuning, we also bring in the Sentences Involving Compositional Knowledge (SICK) dataset consisting of 9,840 additional rows of data \cite{marelli-etal-2014-sick}.

\subsection{Experimental Details and Evaluation Methods}
Before building the Multitask BERT architecture in Figure \ref{fig:multitask_bert}, we conduct a baseline sanity check with minBERT that simply projects from $\texttt{[CLS]}$ embeddings to task predictions using one linear layer per output head, with a naive summing of losses and no gradient modifications.
\paragraph{Multitask BERT} We train the sentiment analysis head with Categorical Cross Entropy loss, the paraphrase detection head with Binary Cross Entropy loss, and the semantic textual similarity head with Mean Square Error loss. As metrics, we track accuracy and Pearson correlation on train and dev data through each epoch. We run experiments according to the optimizations discussed in Section \ref{sec:optimizations}. 

As hyperparameters, we decide to use a batch size of 112 (the maximum before our machines crash), learning rate of $1\times10^{-5}$, dropout probability of 0.5, and weight decay coefficient of $1\times10^{-3}$. We train Multitask BERT for 10-15 epochs.

We also introduce cyclic data loaders, essentially exhausting all the possible data for training. Due to large variance in dataset size across tasks, one zipped batch of data prevents data-abundant tasks from undergoing maximal training. In practice, using a cyclic data loader means that while iterating through the Quora Question Pairs data for paraphrase detection, we exhaust other tasks' data and ensure SST and STS data are passed into training again.

\paragraph{AC-GAN-BERT}
We run a series of experiments to investigate AC-GAN-BERT, training the model as specified in Section \ref{sec:gantraining}:
\begin{itemize}
    \item We test whether there is marginal benefit to using a conditional GAN (conditioned on class labels) vs. normal GAN as in Croce et al. (2020). We accomplish this both quantitatively through test accuracy and qualitatively through visual deciphering of embedding ``quality.''
    \item We run experiments varying the depth of $\mathcal{D}$ and $\mathcal{G}$, examining the relationship between the number of hidden layers of AC-GAN-BERT and model performance.
    \item We also conduct a sensitivity analysis on semi-supervised learning on the sentiment analysis task with AC-GAN-BERT, varying the proportion of labeled data to investigate how resistant the model is to unlabeled data. We accomplish this by using a random seed to mask out increasing numbers of phrases in the SST dataset; these masked out phrases become unlabeled data for the discriminator to classify as real or fake. We seek to quantify a point at which performance tapers off when using unlabeled data with respect to test accuracy.
\end{itemize}
We use a learning rate of $5\times10^{-5}$ and a 100-length noice vector as generator latent space input. Because of computational resource limitations, we train AC-GAN-BERT for 5 epochs, and for the paraphrase detection task train data is truncated to 10,000 sentence pairs and test data to 2,000 sentence pairs. As evaluation methods, we quantitatively assess accuracy scores and qualitatively assess embedding representations.

\subsection{Results}
\label{sec:results}
\paragraph{Multitask BERT}
We list performance metrics for our baseline, several iterations of Multitask BERT, and our final test leaderboard submission in Table \ref{tab:acc}. Some notable optimizations result in large performance boosts, such as tokenization of both sentences separated with $\texttt{[SEP]}$ for semantic textual similarity as opposed to using the same triplet network approach as in paraphrase detection (Light Multitask BERT and beyond have this modification). Interestingly and unexpectedly, adding more data and larger, deeper layers did not positively affect sentiment classification accuracy.  
\renewcommand{\arraystretch}{1.15}
\begin{table}[hbt!]
\centering
\begin{tabular}{|lcccc|}
\hline
Model Details   & Sentiment    & Paraphrase   & Similarity   & \textbf{Overall} \\ \hline \hline
Baseline & 0.410 & 0.532 & 0.102 & 0.498 \\ 
Baseline with PCGrad & 0.437 & 0.565 & 0.211 & 0.536 \\
Pairwise Loss Addition with PCGrad & 0.481 & 0.723 & 0.349 & 0.626 \\
Double Triplet for Paraphrase and STS & \textbf{0.513} & 0.756 & 0.361 & 0.650 
\\
Light Multitask BERT & 0.496 & 0.870 & 0.825 & 0.760  \\ 
Light Multitask BERT Data-Enhanced & 0.502 & 0.878 & 0.822 & 0.764 \\
Multitask BERT Data-Enhanced & 0.501 & \textbf{0.888} & \textbf{0.872} & \textbf{0.775} \\\hline\hline
\textbf{Final Multitask BERT on Test Data} & \textbf{0.516} & \textbf{0.886} & \textbf{0.864} & \textbf{0.778} \\
\hline
\end{tabular}
\caption{Compilation of Multitask BERT Results}
\label{tab:acc}
\end{table}

\paragraph{AC-GAN-BERT} We list classification accuracies of GAN-BERT and AC-GAN-BERT in Table \ref{tab:ganvscgan}. There is no statistically significant boost in accuracy resulting from the inclusion of a conditional generator into the GAN-BERT framework. However, we observe qualitative improvements which are discussed in Section \ref{sec:analysis}.
\renewcommand{\arraystretch}{1.15}
\begin{table}[hbt!]
\centering
\begin{tabular}{|lcc|}
\hline
Model Details   & Sentiment    & Paraphrase \\ \hline \hline
GAN-BERT & 0.502          & 0.816 \\
AC-GAN-BERT Depth 1 & 0.503          & 0.814  \\
AC-GAN-BERT Depth 2 & \textbf{0.510}          & \textbf{0.820}  \\
AC-GAN-BERT Depth 3 & 0.495          & 0.817  \\
\hline
\end{tabular}
\caption{GAN-BERT Performance - Varying Hidden Depth}
\label{tab:ganvscgan}
\end{table}

\begin{figure}[hbt!]
  \begin{minipage}[b]{.52\linewidth}
    \centering
    \hspace{-7mm}
    \includegraphics[scale=0.45]{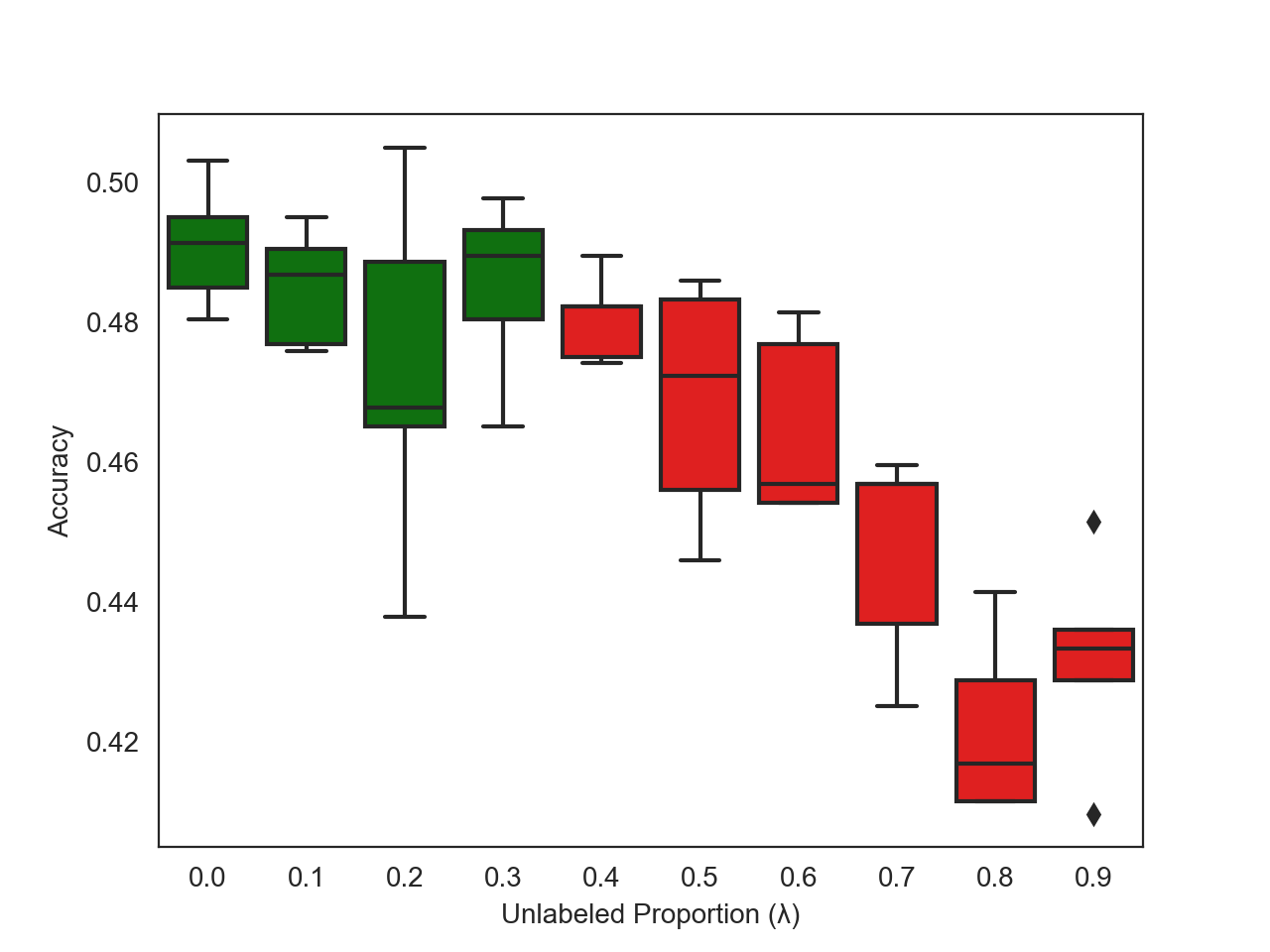}
    \captionof{figure}{Unlabeling-Induced Accuracy Decrease}
    \label{fig:sensitivity}
  \end{minipage}
  \begin{minipage}[b]{.45\linewidth}
    \renewcommand{\arraystretch}{1.15}
    \centering
    \begin{tabular}{|lccc|}
        \hline
        $\lambda$  & Max Acc. & Min Acc. & Mean Acc. \\ \hline \hline
        0.0 & 0.503  & \textbf{0.480} & \textbf{0.491} \\
        0.1 & 0.495  & 0.476 & 0.485 \\
        0.2 & \textbf{0.505}  & 0.437 & 0.473 \\
        0.3 & 0.498  & 0.465 & 0.485 \\
        0.4 & 0.490  & 0.474 & 0.481 \\
        0.5 & 0.486  & 0.446 & 0.469 \\
        0.6 & 0.481  & 0.454 & 0.465 \\
        0.7 & 0.460  & 0.425 & 0.447 \\
        0.8 & 0.441  & 0.411 & 0.422 \\
        0.9 & 0.451  & 0.410 & 0.432 \\
        \hline
        \end{tabular}
    \captionof{table}{Sensitivity Analysis Results}
    \label{tab:sensitivity}
  \end{minipage}
\end{figure}

Figure \ref{fig:sensitivity} and Table \ref{tab:sensitivity} display the results of the semi-supervised learning sensitivity analysis on the SST dataset, where $\lambda$ represents the proportion of the SST dataset that is masked (where individual examples are selected with a random seed). The overall trend is that a larger fraction of unlabeled training data causes AC-GAN-BERT to perform with a lower accuracy. In particular, relative to the performance of AC-GAN-BERT on a fully labeled dataset ($\lambda=0.0$), the performance drop becomes statistically significant ($p<0.05$) when $\lambda=0.4$, hence the color-coding of Figure \ref{fig:sensitivity}. We calculate \textit{p}-values using a one-tailed t-test for difference in sample means, where a sample consists of validation accuracies from all epochs. Table \ref{tab:pvalues} contains all \textit{p}-values in this analysis.

\section{Analysis}
\label{sec:analysis}
\subsection{Multitask BERT}
\paragraph{Discussion} When devising our final architecture on limited compute and a low batch size of 4 or 8 examples, we observe that using a lower learning rate of $8\times10^{-6}$ aided learning, as smaller updates are conducive to better convergence, especially given that fine-tuning is a learning rate sensitive task \cite{li2020rethinking}. We also observe that doubling the number of hidden units in the \textit{shared} and \textit{dense} blocks in Figure \ref{fig:multitask_bert} boosts performance. However, after obtaining more computational power, we increase batch size to 112, the maximum our GPU can handle; we also increase learning rate to $1\times10^{-5}$, as with more examples per batch, we reason that larger update steps can be taken. In general, though we apply dropout regularization, weight decay, and layer normalization, we still observe inevitable overfitting, with training data accuracies very quickly reaching nearly 100\%.
\paragraph{Noteworthy Optimizations} We observe large improvements in paraphrase detection accuracy with the triplet network architecture. Inspired by this, we originally attempt to use the same architecture for STS prediction, but realize that Pearson correlation plateaus at around 0.4. We then switch our STS tokenization mechanism, tokenizing sentence pairs simultaneously with the $\texttt{[SEP]}$ token as a connector. This adjustment improves STS Pearson correlation significantly by around 0.3. Additionally, data augmentation through the SICK and MTEB datasets boosts performance on both paraphrase detection and semantic similarity prediction, but interestingly does not increase sentiment classification accuracy. 

\subsection{AC-GAN-BERT}
Given that GANs are susceptible to mode collapse, we are especially interested in evaluating the \textit{quality} of Generator output \cite{bau2019seeing}. Recall that one of our main motivations for developing a generative adversarial framework is to assess the ability of the Generator to produce high-quality embeddings, in other words, fulfill the same role as BERT. However, ``quality'' can be defined many ways: we know that a good generator's samples will mimic the true data manifold, and in computer vision applications, mere visual inspection can be used to evaluate quality \cite{shmelkov2018good}\cite{salehi2020generative}. In our work, we evaluate generator quality through visual inspection after mathematical manipulation. To qualitatively assess embeddings, we apply t-distributed stochastic neighbor embedding (t-SNE) to reduce hidden dimensionality from 768 (the hidden dimension of $\texttt{[CLS]}$) to 2, allowing embeddings to be plotted on an xy-coordinate plane \cite{van2008visualizing}. 

\begin{figure}[hbt!]
     \centering
     \begin{subfigure}[b]{0.47\textwidth}
         \centering
         \includegraphics[scale=0.08]{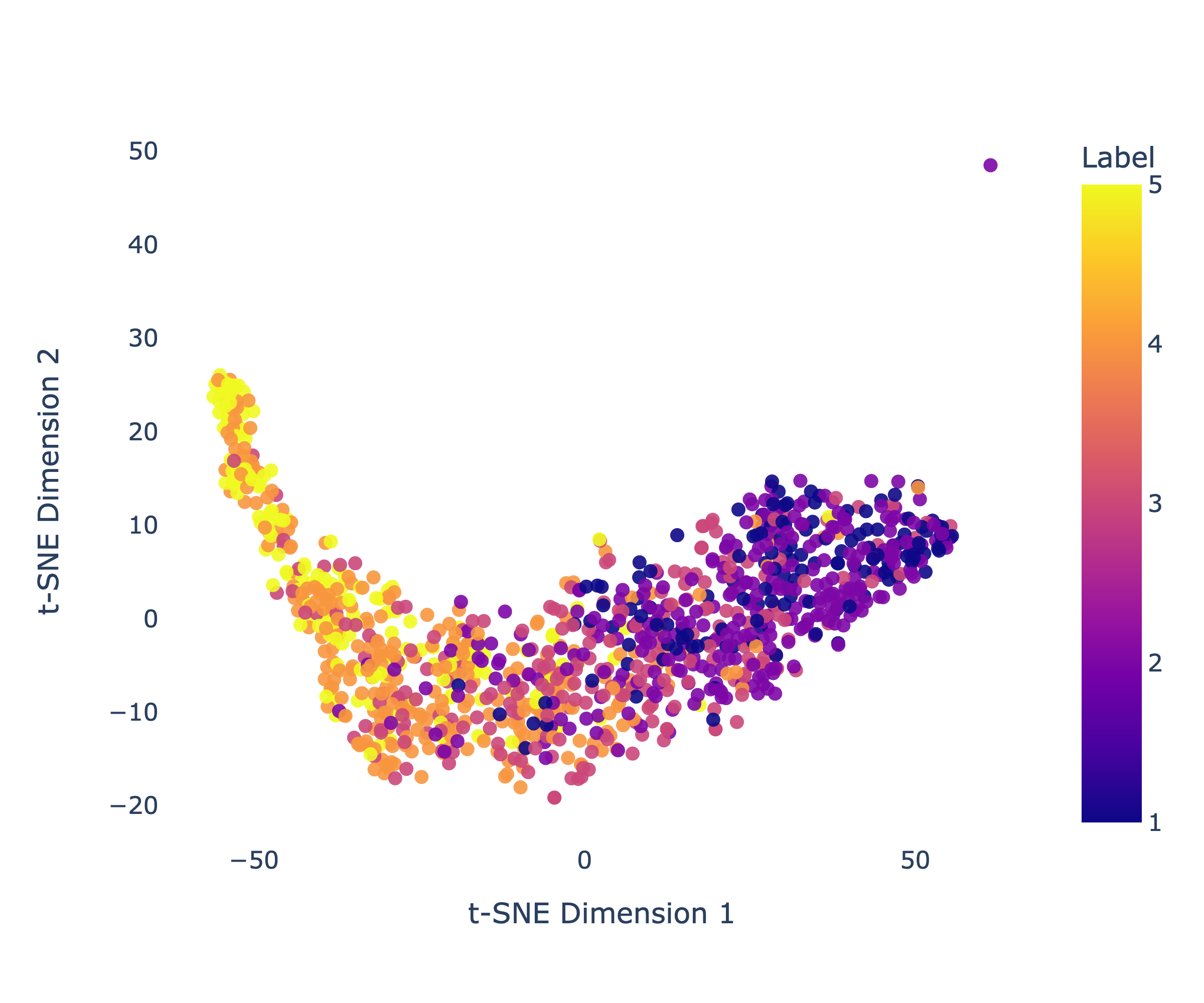}
         \caption{BERT Embeddings}
     \end{subfigure}
     \hfill
     \begin{subfigure}[b]{0.47\textwidth}
         \centering
         \includegraphics[scale=0.08]{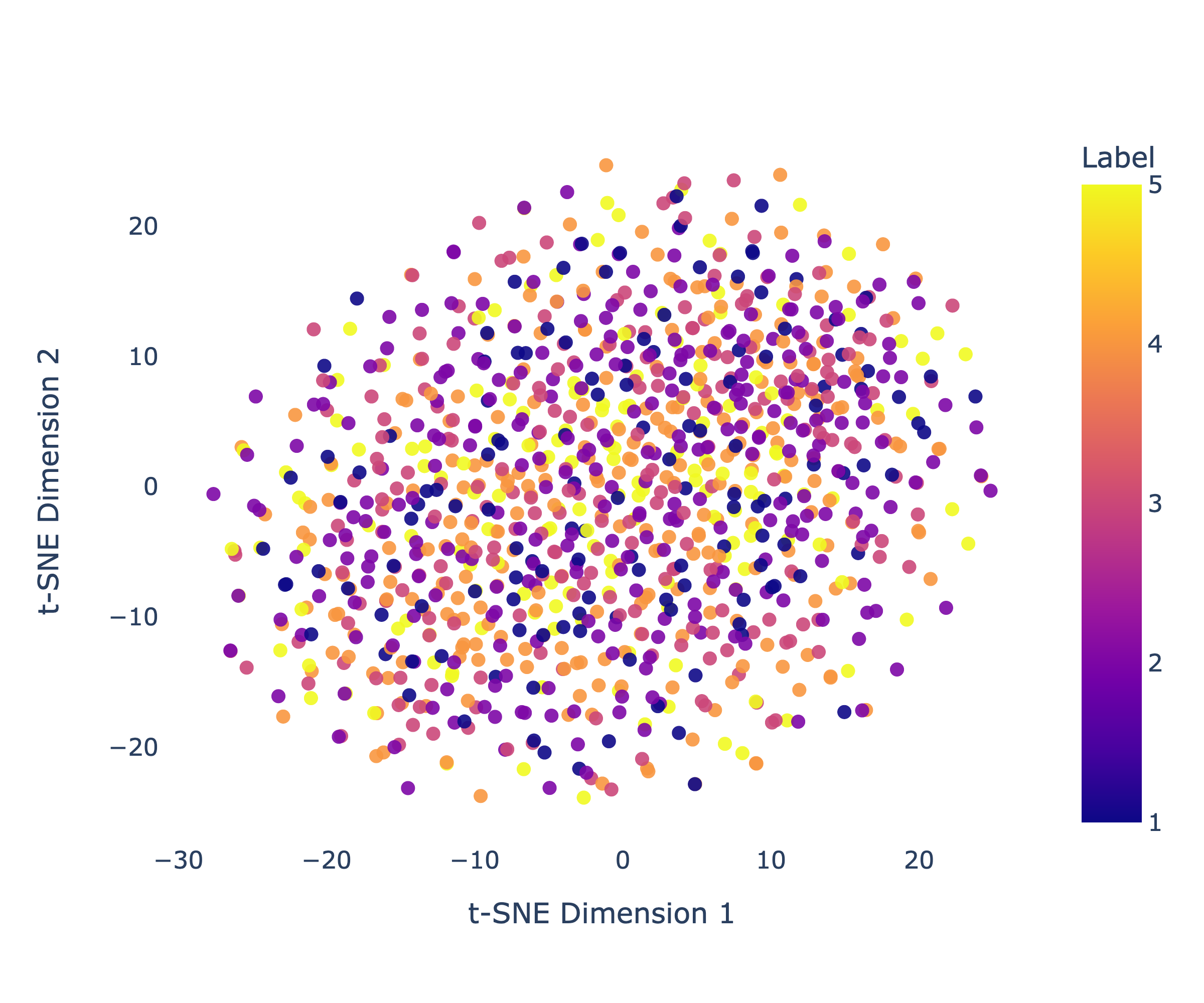}
         \caption{Generator Embeddings}
     \end{subfigure}
    \caption{Unconditional GAN-BERT - 40\% Unlabeled}
    \label{fig:embed_gan40}
\end{figure}

\begin{figure}[hbt!]
     \centering
     \begin{subfigure}[b]{0.47\textwidth}
         \centering
         \includegraphics[scale=0.08]{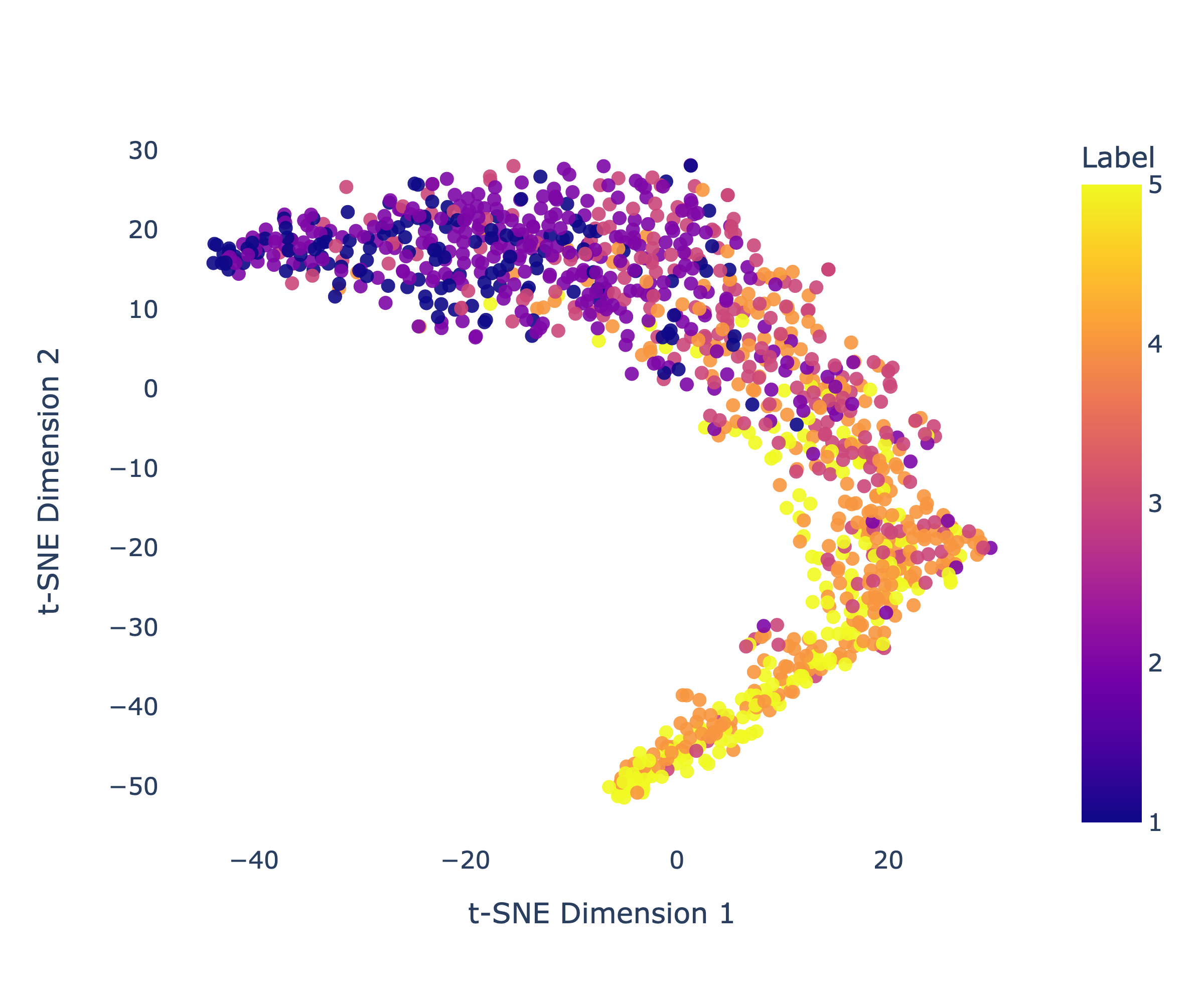}
         \caption{BERT Embeddings}
     \end{subfigure}
     \hfill
     \begin{subfigure}[b]{0.47\textwidth}
         \centering
         \includegraphics[scale=0.08]{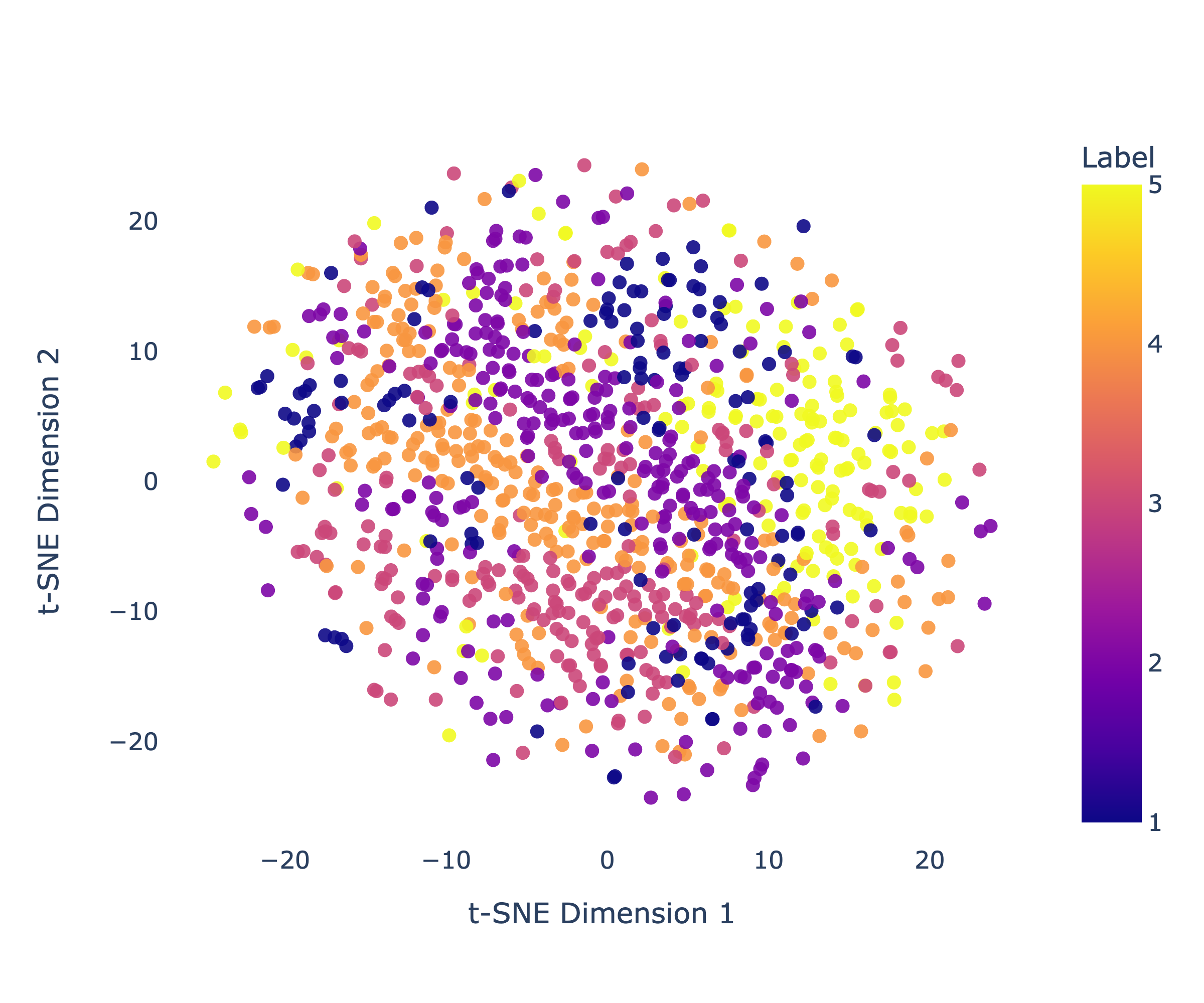}
         \caption{Generator Embeddings}
     \end{subfigure}
    \caption{AC-GAN-BERT - 40\% Unlabeled}
    \label{fig:embed_cgan40}
\end{figure}

\begin{figure}[hbt!]
     \centering
     \begin{subfigure}[b]{0.47\textwidth}
         \centering
         \includegraphics[scale=0.08]{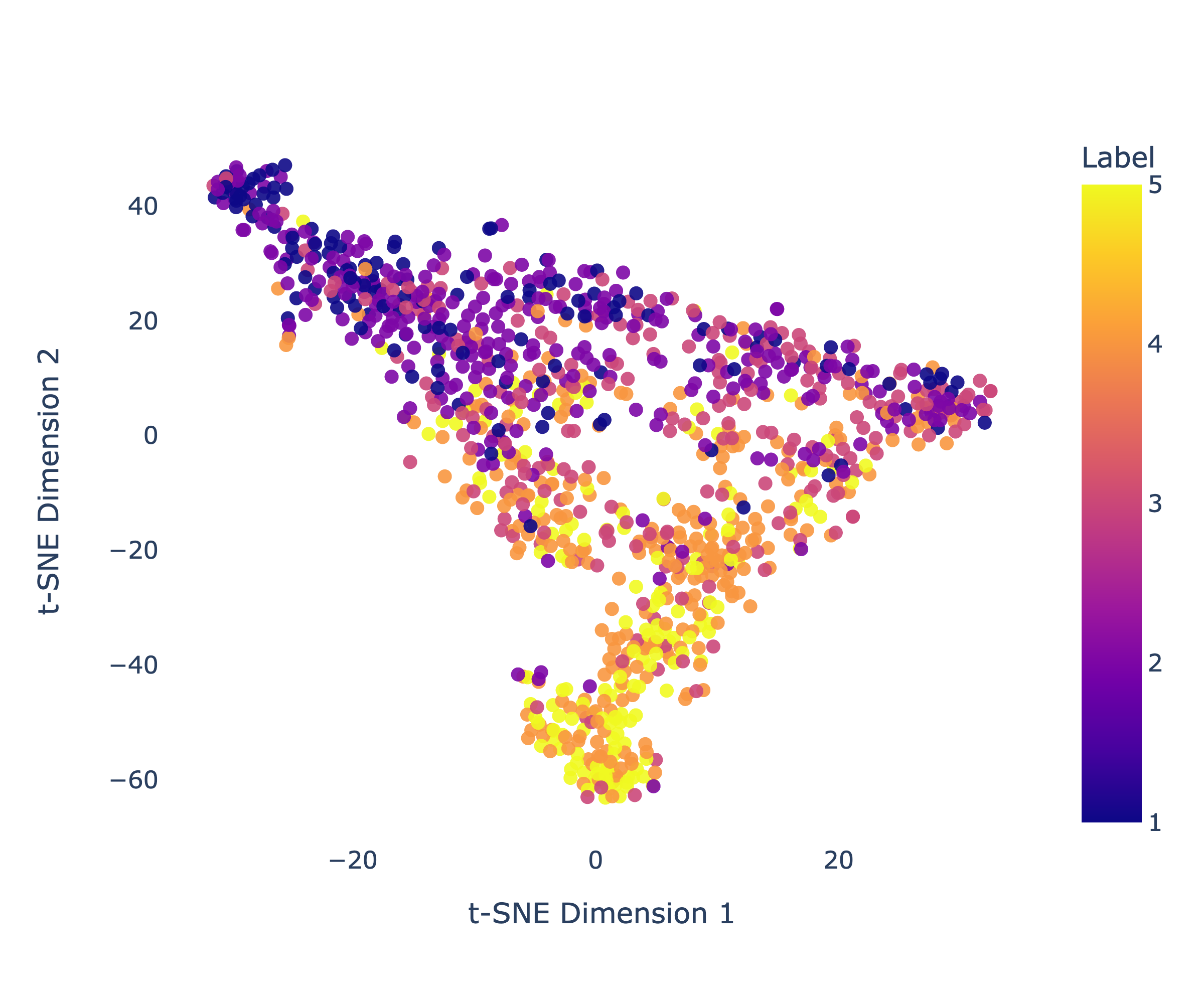}
         \caption{BERT Embeddings}
     \end{subfigure}
     \hfill
     \begin{subfigure}[b]{0.47\textwidth}
         \centering
         \includegraphics[scale=0.08]{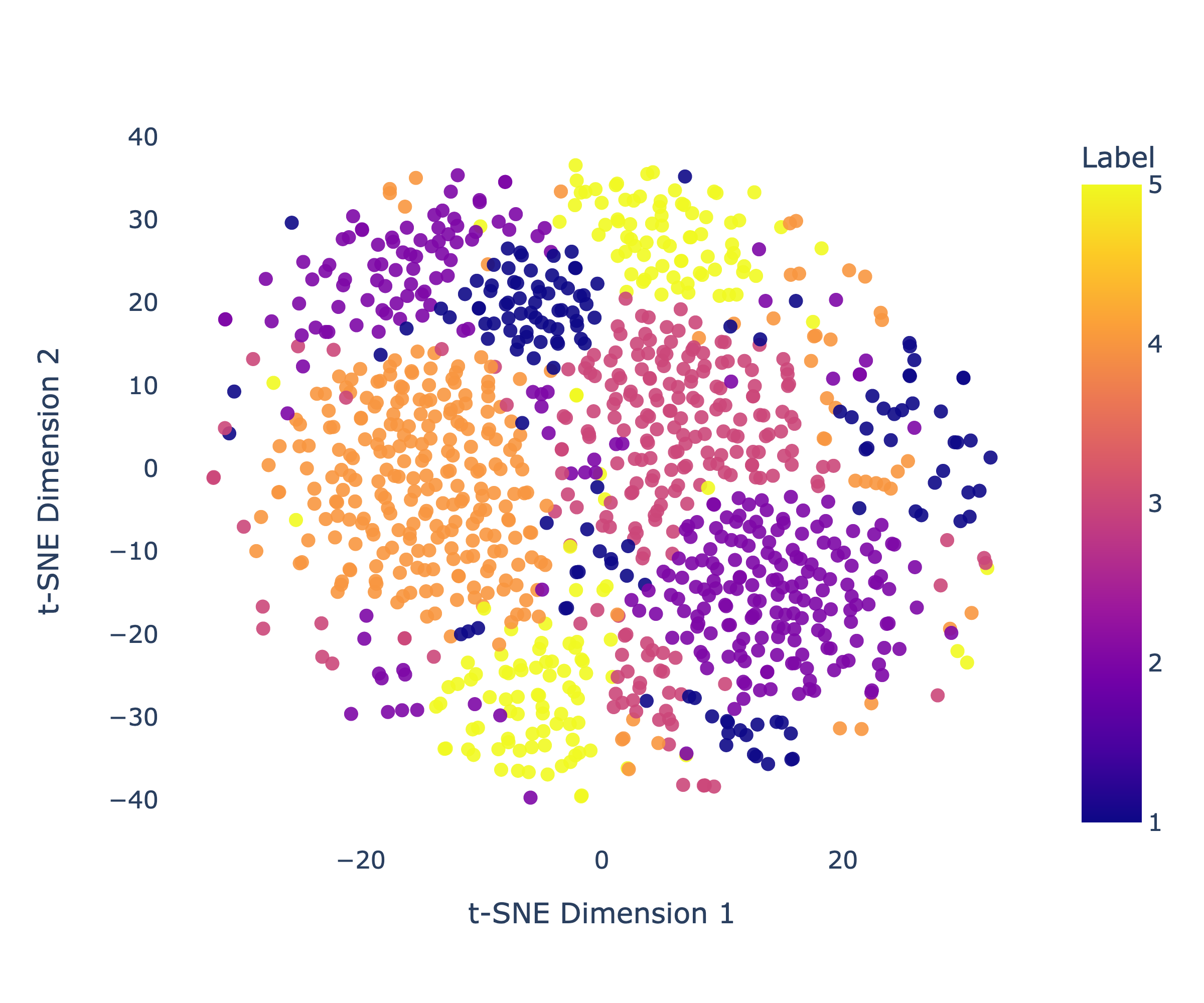}
         \caption{Generator Embeddings}
     \end{subfigure}
    \caption{AC-GAN-BERT - 80\% Unlabeled}
    \label{fig:embed_cgan80}
\end{figure}

We extract both BERT and Generator embeddings for three models: \textbf{1}) Croce et al.'s unconditional GAN-BERT trained with 40\% unlabeled data, \textbf{2}) AC-GAN-BERT trained with 40\% unlabeled data, and \textbf{3}) AC-GAN-BERT trained with 80\% unlabeled data. Comparison between the first and second allows us to isolate the marginal benefit in embedding quality of a conditional generator, and comparison between the second and third reveals the effect of highly-skewed datasets on embedding quality. Figure \ref{fig:embed_gan40} illustrates t-SNE results of Model \textbf{1} embeddings, Figure \ref{fig:embed_cgan40} illustrates t-SNE results of Model \textbf{2} embeddings, and Figure \ref{fig:embed_cgan80} illustrates t-SNE results of Model \textbf{3} embeddings.

Critically, we notice from Figure \ref{fig:embed_gan40}(b) that when using an unconditional Generator, there is \textit{no spatial relationship} between embedding and class label. However, we notice from Figure \ref{fig:embed_cgan40}(b) that embeddings seem to be clustered according to their class label, with equally-labeled embeddings residing in similar spatial regions. This observation is even more stark in Figure \ref{fig:embed_cgan80}(b), which reveals that embedding regions are well-defined according to their class label. In all cases, BERT embeddings possess a spatial relationship with class label, which of course is a validation of BERT's quality. But similar observations in conditional Generator embeddings serve as promising validation of their diversity and quality. In general, Generators can fail to map from latent space to their full output domain, instead creating outputs with very little variation. If this was the case in $\mathbb{R}^{768}$, then projecting embeddings down to $\mathbb{R}^2$ with t-SNE would surely cause mode collapse to manifest visually. But because we do not observe this, we can fairly confidently conclude an avoidance of mode collapse. 

\section{Conclusion}
In this study, we present Multitask BERT, a novel architecture optimized for multitask fine-tuning on sentiment classification, paraphrase detection, and semantic textual similarity prediction. As a result of data augmentation, loss pairing, and custom tokenization, we achieve an overall performance score of 0.778 on test data. Additionally, we introduce AC-GAN-BERT, which incorporates generative adversarial learning to enhance BERT's capabilities. The conditional generator within AC-GAN-BERT successfully produces class-conditioned embeddings, avoiding mode collapse and showing clear spatial correlation with class labels. Our results call for future work on more advanced loss functions and better regularization mechanisms to further maximize accuracy. Our conditional generator also offers potential as a student model for BERT knowledge distillation.

\bibliographystyle{unsrt}
\bibliography{references}
\clearpage

\appendix
\section*{Appendix}

\renewcommand{\arraystretch}{1.25}
\begin{table}[hbt!]
\centering
\footnotesize
\begin{tabular}{|c||cccccccccc|}
\hline
$\lambda$ & \multicolumn{10}{c|}{Columnwise \textit{p}-values Relative to $\lambda=\frac{\text{column number}}{10}$} \\ \hline\hline
0.0             & \multicolumn{1}{c|}{0.500}   & \multicolumn{1}{c|}{}        & \multicolumn{1}{c|}{}      & \multicolumn{1}{c|}{}        & \multicolumn{1}{c|}{}        & \multicolumn{1}{c|}{}        & \multicolumn{1}{c|}{}        & \multicolumn{1}{c|}{}      & \multicolumn{1}{c|}{}      &       \\ \hline
0.1             & \multicolumn{1}{c|}{0.152}   & \multicolumn{1}{c|}{0.500}   & \multicolumn{1}{c|}{}      & \multicolumn{1}{c|}{}        & \multicolumn{1}{c|}{}        & \multicolumn{1}{c|}{}        & \multicolumn{1}{c|}{}        & \multicolumn{1}{c|}{}      & \multicolumn{1}{c|}{}      &       \\ \hline
0.2             & \multicolumn{1}{c|}{0.085}   & \multicolumn{1}{c|}{0.170}   & \multicolumn{1}{c|}{0.500} & \multicolumn{1}{c|}{}        & \multicolumn{1}{c|}{}        & \multicolumn{1}{c|}{}        & \multicolumn{1}{c|}{}        & \multicolumn{1}{c|}{}      & \multicolumn{1}{c|}{}      &       \\ \hline
0.3             & \multicolumn{1}{c|}{0.215}   & \multicolumn{1}{c|}{0.510}   & \multicolumn{1}{c|}{0.819} & \multicolumn{1}{c|}{0.500}   & \multicolumn{1}{c|}{}        & \multicolumn{1}{c|}{}        & \multicolumn{1}{c|}{}        & \multicolumn{1}{c|}{}      & \multicolumn{1}{c|}{}      &       \\ \hline
0.4             & \multicolumn{1}{c|}{0.033}   & \multicolumn{1}{c|}{0.190}   & \multicolumn{1}{c|}{0.738} & \multicolumn{1}{c|}{0.250}   & \multicolumn{1}{c|}{0.500}   & \multicolumn{1}{c|}{}        & \multicolumn{1}{c|}{}        & \multicolumn{1}{c|}{}      & \multicolumn{1}{c|}{}      &       \\ \hline
0.5             & \multicolumn{1}{c|}{0.017}   & \multicolumn{1}{c|}{0.047}   & \multicolumn{1}{c|}{0.385} & \multicolumn{1}{c|}{0.063}   & \multicolumn{1}{c|}{0.092}   & \multicolumn{1}{c|}{0.500}   & \multicolumn{1}{c|}{}        & \multicolumn{1}{c|}{}      & \multicolumn{1}{c|}{}      &       \\ \hline
0.6             & \multicolumn{1}{c|}{<0.001} & \multicolumn{1}{c|}{0.010}   & \multicolumn{1}{c|}{0.271} & \multicolumn{1}{c|}{0.019}   & \multicolumn{1}{c|}{0.021}   & \multicolumn{1}{c|}{0.347}   & \multicolumn{1}{c|}{0.500}   & \multicolumn{1}{c|}{}      & \multicolumn{1}{c|}{}      &       \\ \hline
0.7             & \multicolumn{1}{c|}{<0.001} & \multicolumn{1}{c|}{<0.001} & \multicolumn{1}{c|}{0.044} & \multicolumn{1}{c|}{<0.001} & \multicolumn{1}{c|}{<0.001} & \multicolumn{1}{c|}{0.035}   & \multicolumn{1}{c|}{0.044}   & \multicolumn{1}{c|}{0.500} & \multicolumn{1}{c|}{}      &       \\ \hline
0.8             & \multicolumn{1}{c|}{<0.001} & \multicolumn{1}{c|}{<0.001} & \multicolumn{1}{c|}{0.002} & \multicolumn{1}{c|}{<0.001} & \multicolumn{1}{c|}{<0.001} & \multicolumn{1}{c|}{<0.001} & \multicolumn{1}{c|}{<0.001} & \multicolumn{1}{c|}{0.012} & \multicolumn{1}{c|}{0.500} &       \\ \hline
0.9             & \multicolumn{1}{c|}{<0.001} & \multicolumn{1}{c|}{<0.001} & \multicolumn{1}{c|}{0.007} & \multicolumn{1}{c|}{<0.001} & \multicolumn{1}{c|}{<0.001} & \multicolumn{1}{c|}{<0.001} & \multicolumn{1}{c|}{<0.001} & \multicolumn{1}{c|}{0.075} & \multicolumn{1}{c|}{0.849} & 0.500 \\ \hline
\end{tabular}
\caption{Summary of \textit{p}-values Relative to Unlabeled Percentages}
\label{tab:pvalues}
\end{table}
The way to interpret this table is as follows: in column $i$, the entry at row $i$ will always equal 0.5, because the \textit{p}-value resulting from a one-tailed t-test of difference in sample means, where both samples are the same, will always equal 0.5. In column $i$, the entry in row $i+j$, where $j$ ranges from 1 to $9-i$, is the \textit{p}-value representing the significance of the decrease in validation accuracy between training AC-GAN-BERT with $\lambda=\frac{i}{10}$ and $\lambda=\frac{i+j}{10}$.

\end{document}